\documentclass[letterpaper]{article} 
\usepackage{aaai2027}

\usepackage{xcolor}
\usepackage{hyperref}
\definecolor{myorange}{RGB}{245,156,74}
\definecolor{mygray}{gray}{0.4}        
\definecolor{myred}{RGB}{197,41,114}
\hypersetup{
  colorlinks=true,
  urlcolor=blue,
  linkcolor=red,
  citecolor=-myorange,
}

\makeatletter
\DeclareRobustCommand\onedot{\futurelet\@let@token\@onedot}
\def\@onedot{\ifx\@let@token.\else.\null\fi\xspace}

\def\eg{\emph{e.g}\onedot} 

\def\ie{\emph{i.e}\onedot}

\makeatother

\usepackage{graphicx} 
\usepackage{natbib}  
\usepackage{caption} 
\usepackage{algorithm}
\usepackage{algorithmic}
\usepackage{xspace}
\usepackage{amsfonts}
\usepackage{multirow}
\usepackage{graphicx}
\usepackage{booktabs}
\usepackage{amsmath}
\usepackage{diagbox}
\usepackage{makecell}
\usepackage{utfsym}
\usepackage{subcaption}

\definecolor{darkgreen}{RGB}{0,200,0}
\definecolor{darkred}{RGB}{200,0,0}

\usepackage[nameinlink,capitalize]{cleveref}

\usepackage{newfloat}
\usepackage{listings}
\DeclareCaptionStyle{ruled}{labelfont=normalfont,labelsep=colon,strut=off} 
\floatstyle{ruled}
\newfloat{listing}{tb}{lst}{}
\floatname{listing}{Listing}

\usepackage{booktabs}

\nocopyright 

\title{GS-RealBlur: A Flexible Data Acquisition Framework for \\  Real-World Image Deblurring}
\author{
    Mingyang Chen,
    Zhilu Zhang,
    Honglei Xu,
    Renlong Wu,
    Xiaohe Wu,
    Wangmeng Zuo
}
\affiliations{
    Harbin Institute of Technology, Harbin, China\\
    youngmchan269@gmail.com, cszlzhang@outlook.com, \{cshongleixu, hirenlongwu, csxhwu, cswmzuo\}@gmail.com \\
    \vspace{2mm}
    Project: \textcolor{red}{https://youngmchan.github.io/GS-RealBlur/} 
}

\begin{document}

\pagestyle{plain}
\setcounter{page}{1}

\maketitle

\begin{abstract}
High-quality, large-scale paired data is essential for training learning-based image deblurring models. However, synthetic blurry images generally lack realism, while real-world captured images require complex and inflexible camera systems. In this work, we propose GS-RealBlur, a data acquisition framework for real-world image deblurring, achieving both blur realism and acquisition flexibility. Specifically, we use a handheld camera to capture blurry images, and deploy a gimbal to densely capture sharp images of the same scene. We reconstruct the 3D representation of sharp images and calibrate the camera pose of each blurry frame within this 3D. The image rendered from this 3D according to the pose serves as the sharp counterpart. To better align the rendered image with the blurry image, we introduce a Blur-aware Pose Refinement (BPR) module that refines the pose using appearance consistency and centroid alignment constraints. Leveraging GS-RealBlur, we construct a high-quality and diverse dataset. Extensive experiments demonstrate that a deblurring model trained on our dataset achieves superior generalization performance across various real-world deblurring benchmarks, consistently outperforming models trained on existing synthetic and real-world datasets. The code and dataset will be made publicly available.
\end{abstract}

\section{Introduction}
\label{sec:intro}

As a fundamental challenge in image restoration~\cite{su2022survey,zhai2023comprehensive,chen2021pre,zhang2021learning,lin2024unirestorer}, image motion deblurring~\cite{chen2022simple,zamir2022restormer,wang2022uformer, liang2021swinir, tsai2022stripformer,tao2018scale,park2020multi,gao2019dynamic,zhou2019spatio} aims to reconstruct a latent sharp image from a blurry one degraded by motion from the camera and dynamic objects. To address this, learning-based methods~\cite{nah2017deep, nah2019ntire, su2017deep, zhou2019davanet, shen2019human, deng2021multi, li2021arvo,brooks2019learning, zhang2020deblurring} leverage paired datasets comprising blurry and sharp images, enabling supervised training of deblurring models. As a result, the performance and generalization capability of these models are directly contingent upon the quality and scale of the training data. Datasets of higher quality and larger scale are conducive to the development of more robust and effective deblurring models.

\begin{figure*}[t]
    \centering
    \includegraphics[width=0.97\textwidth]{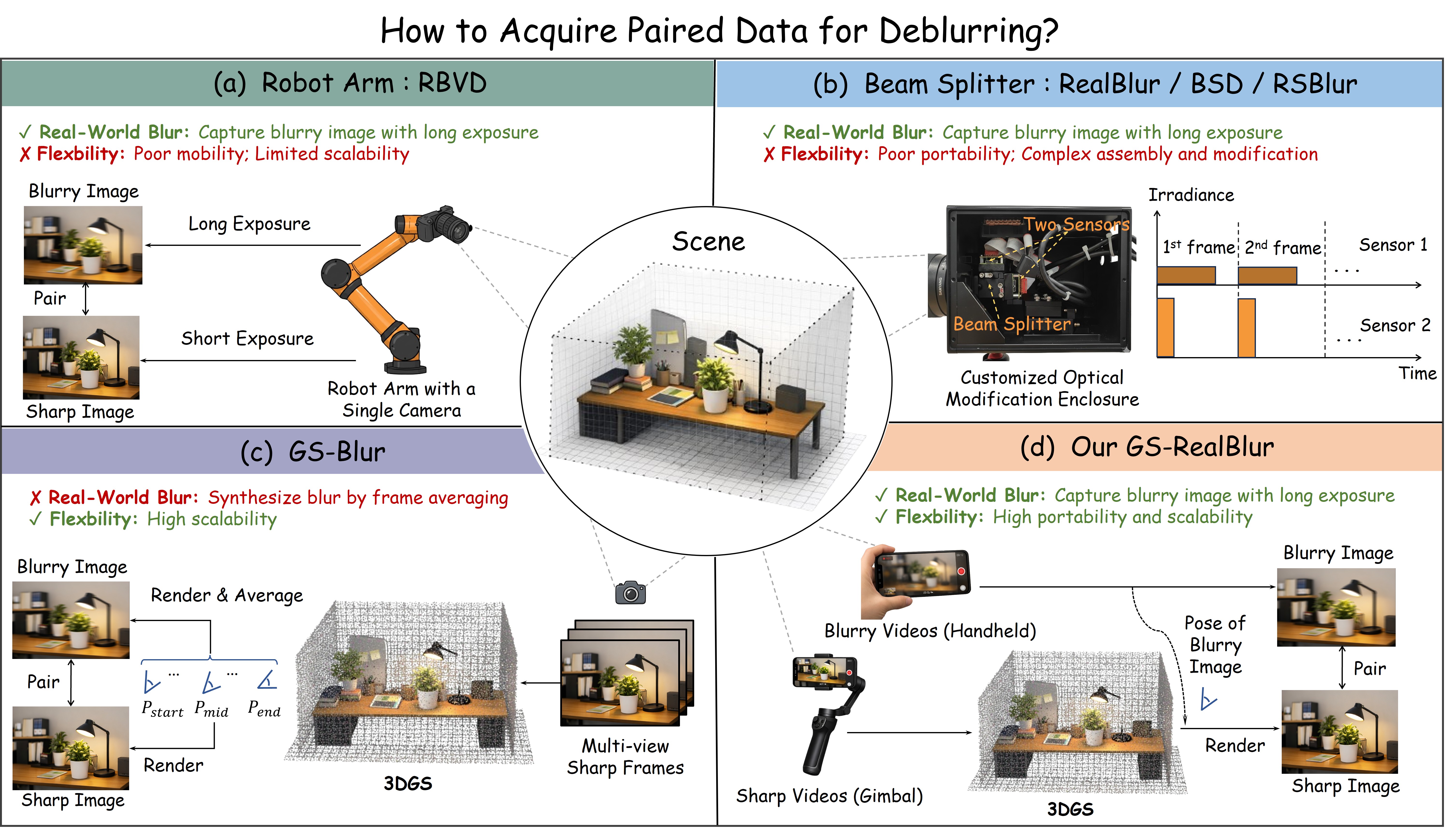}
    \vspace{-1mm}
    \caption{\textbf{Comparison of paired data acquisition methods for deblurring.}
    Existing methods either capture real-world blur with restricted flexibility
    (\eg, (a) Robot arm~\cite{zhu2022deep} and
    (b) Beam-splitter~\cite{rim2020real,rim2022realistic,zhong2020efficient})
    or synthesize unrealistic blur
    (\eg, (c) GS-Blur~\cite{lee2024gs}).
    In contrast, our GS-RealBlur (d) achieves both blur realism and acquisition
    flexibility by pairing real-world blurry images from consumer-grade devices
    with sharp correspondences rendered from 3DGS~\cite{kerbl20233d}.}
    \label{fig:limitation}
    \vspace{-2mm}
\end{figure*}

However, obtaining a large number of high-quality blurry-sharp image pairs is not trivial.
A straightforward strategy is to aggregate multiple consecutive frames to synthesize blurry images. Early works~\cite{nah2017deep, nah2019ntire, su2017deep, zhou2019davanet, shen2019human, deng2021multi, li2021arvo} employ high-speed cameras to capture sharp videos for blur synthesis. Recently, GS-Blur~\cite{lee2024gs} (see Figure~\ref{fig:limitation}(c)) reconstructs multi-view images into a 3D representation using 3D Gaussian Splatting~\cite{kerbl20233d}, and then renders much denser images along randomly generated motion trajectories, averaging them to produce more realistic blurry images.
Nevertheless, the blur synthesized by such methods still exhibits a distribution gap compared to real-world blur. This discrepancy arises because the discrete aggregation process fundamentally differs from the continuous photoelectric integration in a physical camera. While it is possible to approximate, completely closing this gap remains a significant challenge.

An alternative approach involves capturing real-world blurry-sharp image pairs. The primary challenge in this paradigm is ensuring the alignment of data pairs in terms of color and spatial position. RBVD~\cite{zhu2022deep} addresses this by controlling the position and movement of a camera mounted on a robotic arm to capture aligned short- and long-exposure pairs, as illustrated in Figure~\ref{fig:limitation}(a). Other methods, such as those used to create the RealBlur~\cite{rim2020real}, BSD~\cite{zhong2020efficient} and RSBlur~\cite{rim2022realistic} datasets, employ beam splitters. It can split the beam into two cameras, thereby enabling synchronous capture of short- and long-exposure pairs, as shown in Figure~\ref{fig:limitation}(b). However, these hardware setups are characterized by complex designs and lack portability. The inherent inflexibility of such systems restricts the scale and diversity of the datasets they produced.

It is evident that existing data acquisition methods for deblurring are constrained by a trade-off between blur realism and acquisition flexibility. In this work, we aim to design a framework that simultaneously achieves both objectives. This framework is guided by several key principles: (1) the blurry images must be captured in real-world scenarios, (2) the capture device should be portable and as simple as possible, (3) the image pairs should be aligned while the sharp images should be high-quality.

To this end, we propose GS-RealBlur, a flexible data acquisition framework for real-world image deblurring, as shown in Figure~\ref{fig:limitation}(d). Specifically, we capture videos of a scene using consumer-grade devices (\eg, smartphones) in both handheld and gimbal-mounted configurations. Frames captured with the handheld setup naturally exhibit motion blur, while those captured with the gimbal remain sharp. We then use the sharp frames to reconstruct a 3D representation of the scene and calibrate the camera pose of blurry frames within this 3D. This enables the rendering of a sharp image that is spatially aligned with each blurry frame.
A critical challenge in this pipeline is that directly estimating the poses of blurry images using structure-from-motion tools like GLOMAP~\cite{pan2024global} often yields inaccurate results due to the detrimental effects of blur degradation.
To address this issue, we introduce a Blur-aware Pose Refinement (BPR) module. BPR optimizes the pose by an appearance consistency loss and a centroid alignment regularizer, enabling the rendering to be better aligned with the blurry counterpart.
Specifically, the centroid alignment regularizer serves as a regularizer to mitigate off-center translation misalignment between a blurry and a rendered sharp image. This is achieved by enforcing that the centroid of the estimated blur kernel coincides with its geometric center, ensuring that the rendered sharp frame is modeled as corresponding to the temporal midpoint of the blur formation process.

Finally, we use the data acquisition framework to construct a single-image deblurring dataset consisting of diverse indoor and outdoor scenes, under both daytime and nighttime conditions, with 13,209 blurry-sharp pairs in total. Extensive experiments demonstrate that a deblurring model trained on our dataset exhibits superior generalization performance on real-world scenarios compared to models trained on existing datasets, validating the effectiveness of GS-RealBlur.

The contributions are summarized as follows: 
\begin{itemize}
\item We propose GS-RealBlur, a data acquisition framework for real-world image deblurring, which achieves both blur realism and acquisition flexibility.
\item We propose Blur-aware Pose Refinement (BPR) to optimize the pose of blurry images by appearance consistency and centroid alignment regularizer.
\item We construct a single-image deblurring dataset using GS-RealBlur. Extensive experiments demonstrate that the deblurring model trained on this dataset performs better generalization  than ones trained on existing datasets in real-world scenarios.
\end{itemize}

\section{Related Work}
\label{sec:related works}

\subsection{Image Deblurring Methods}
\label{sec:2.1}

Image deblurring is a long-standing challenge in computer vision.
Traditional methods~\cite{richardson1972bayesian,lucy1974iterative,krishnan2011blind,krishnan2009fast} typically formulate it as a MAP-based optimization problem~\cite{fergus2006removing} under the assumption of a uniform blur kernel. 
However, they often struggle in real-world scenarios where blur is non-uniform and kernels are difficult to estimate.
With the development of deep learning, methods are shifted toward end-to-end restoration models. 
Early CNN-based approaches~\cite{nah2017deep,gao2019dynamic,tao2018scale} 
adopted multi-scale architectures for deblurring. Subsequent studies~\cite{chen2021hinet,zamir2021multi} explored multi-stage restoration frameworks to enhance feature interaction.
More recently, Transformer-based~\cite{zamir2022restormer,wang2022uformer,vaswani2017attention,liang2021swinir,tsai2022stripformer} architectures, further introduced attention mechanisms to model long-range dependencies. 
Despite these increasingly complex designs, recent studies~\cite{chen2022simple,cho2021rethinking} suggest that carefully designed simple baselines, can still achieve competitive performance by optimizing basic block designs.
However, these models still depend on large-scale, high-quality blurry–sharp pairs, which have motivated extensive efforts in data acquisition.

\subsection{Deblurring Datasets}
\label{sec:2.2}
\subsubsection{Synthetic Deblurring Datasets.} Synthetic blur datasets are widely used for training deblurring models. 
Mainstream methods synthesize blur images by averaging high-frame-rate video frames. 
GoPro~\cite{nah2017deep} pioneered this strategy using high-speed sequences captured by a GoPro camera.
Subsequent works extended this paradigm in different settings.  
REDS~\cite{nah2019ntire} extends it to 120 FPS dynamic scenes, while DVD~\cite{su2017deep} adopts a similar pipeline with handheld devices.
Stereo Blur~\cite{zhou2019davanet} applies it to binocular videos to produce stereo pairs.
Except for GoPro, most datasets~\cite{nah2019ntire, su2017deep, zhou2019davanet, shen2019human, deng2021multi} interpolate frames before averaging to accumulation to approximate better exposure blur.
Although scalable, these datasets rely on high-speed cameras that require sufficient illumination(capture with short exposure)~\cite{zhang2025exposure}, limiting coverage of low-light or night-time scenes. 
To avoid this dependency, GS-Blur~\cite{lee2024gs} reconstructs scenes using 3DGS~\cite{kerbl20233d} and synthesizes blur by integrating multi-view renderings along simulated camera trajectories. 
However, these methods are still constrained by the simplified mechanism, resulting in a distribution gap from real camera imaging processes.

In addition, several works move beyond the multi-frame averaging paradigm. Brooks et al.~\cite{brooks2019learning} synthesize blur by predicting spatially varying linear blur kernels from two consecutive sharp frames, but the linear assumption is insufficient. Zhang et al.~\cite{zhang2020deblurring} employ GAN~\cite{goodfellow2020generative} learned from real blurry images to generate blur images from a single sharp input, but the absence of explicit physical imaging modeling limits blur realism. 
\begin{table}[t]
\centering
\caption{\textbf{Comparison of Deblurring Datasets}}
\vspace{-2mm}
\label{table:dataset_comparison}

\scriptsize
\setlength{\tabcolsep}{2.5pt}
\renewcommand{\arraystretch}{1.05}

\resizebox{\columnwidth}{!}{%
\begin{tabular}{@{}lcccc@{}}
\toprule
Dataset & Blur Type & \#Pairs & Exp. Time (ms) & Resolution \\
\midrule
RealBlur
& Real & 4,738 & 500 & 680$\times$772 \\
RBVD
& Real & 2,164 & 16--33 & 1443$\times$960 \\
BSD
& Real & 33,000 & 1--24 & 1280$\times$720 \\
RSBlur
& Real + Synthetic & 13,358 & 100 & 1920$\times$1200 \\
GS-Blur
& Synthetic & 156,209 & -- & -- \\
\textbf{GS-RealBlur}
& Real & 13,209 & 1--100 & 3840$\times$2160 \\
\bottomrule
\end{tabular}
}
\vspace{-4mm}
\end{table}

\subsubsection{Real-world Deblurring Datasets.} 
To improve performance in real scenarios, several works~\cite{rim2020real, rim2022realistic, zhong2020efficient}collect real blurry–sharp pairs using dual-sensor systems with beam splitters, which split incoming light into two paths and capture images with different exposure times. Representative datasets include BSD~\cite{zhong2020efficient}, RealBlur~\cite{rim2020real}, and RSBlur~\cite{rim2022realistic}. Table~\ref{table:dataset_comparison} shows the information about existing deblurring datasets. Although these methods better reflect real blur formation, they require complex hardware and precise calibration, leading to poor flexibility and limited scalability. Differences between sensors may also cause photometric inconsistencies. RBVD~\cite{zhu2022deep} uses a robotic arm to precisely control camera motion, enabling repeatable trajectories and accurate alignment of training pairs. the system has a large physical footprint, requires complex setup, and offers limited mobility, resulting in restricted scene coverage and poor scalability.

Some other works~\cite{su2017deep, zhang2020deblurring, kohler2012recording,xu2025selfhvd} collect blurry images in real-world scenarios, but the lack of reliable aligned sharp images limits their use for supervised training. In general, existing real-world datasets remain limitation due to hardware constraints and capture difficulty.

\begin{figure*}[t!]
    \centering
    \includegraphics[width=1\linewidth]{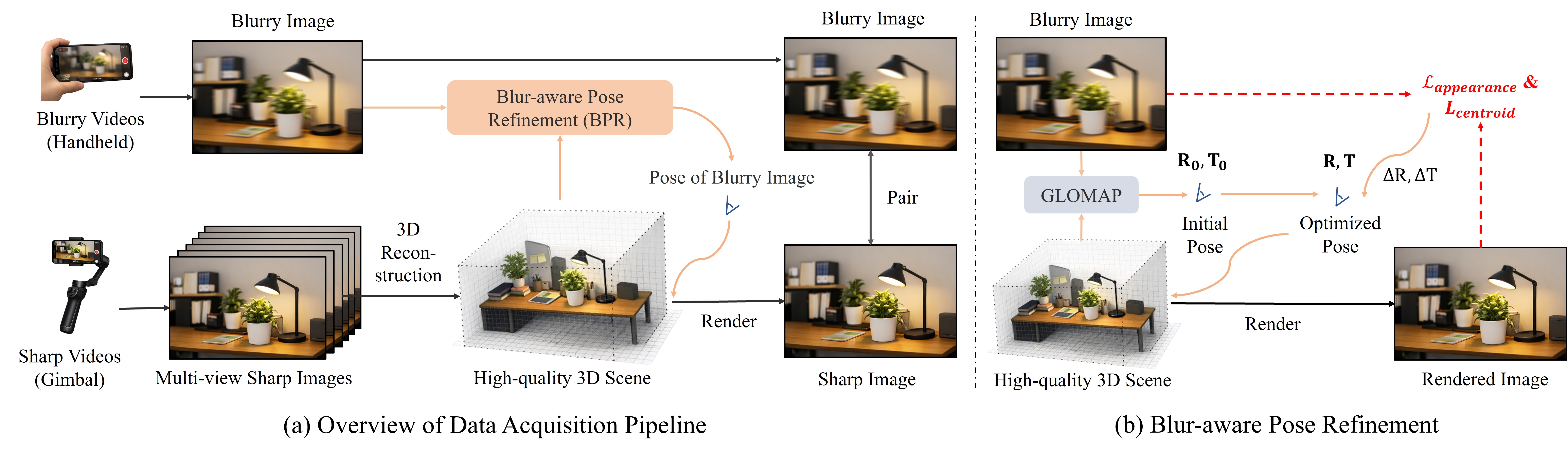}
    \vspace{-4mm}
    \caption{\textbf{Overview of GS-RealBlur}. We first capture real-world blurry images with handheld consumer-grade devices. Then we densely capture multi-view sharp images with a gimbal for high-quality 3D scene reconstruction. Given the camera pose of a blurry image optimized by Blur-aware Pose Refinement (BPR) module, the rendered image from this 3D is taken as the sharp counterpart. }
    \label{fig:pipeline}
    \vspace{-2mm}
\end{figure*}

\section{Method}
\label{sec:method}

\subsection{Motivation}
\label{sec:3.1}
Existing data acquisition methods either synthesize blurry images or use sophisticated camera systems to capture real-world blurry images. They are constrained by a trade-off between blur realism and acquisition flexibility. 
\subsubsection{Blur Realism.} 
During exposure, the camera sensor integrates the incoming light over time to form an image. If the camera moves during exposure, motion blur occurs.
To simulate this process for synthesizing blurred images, 
synthetic methods~\cite{nah2017deep, nah2019ntire, su2017deep,zhou2019davanet,shen2019human,deng2021multi,li2021arvo} use high-speed cameras to capture sharp videos and aggregate consecutive sharp frames, but the discrete temporal sampling still introduces a gap from real blur.
GS-Blur~\cite{lee2024gs} instead reconstructs scenes with 3DGS~\cite{kerbl20233d} and averages much denser renderings images along randomly generated motion trajectories, producing more realistic blurry images. 
Nevertheless, such synthetic blur still exhibits a distribution gap to real-world blur in three main aspects~\cite{rim2022realistic}. 
First, discrete frame averaging cannot fully reproduce the continuous photoelectric integration of physical cameras.
Second, synthetic blur is usually generated in sRGB or linear RGB space, while real blur forms in the RAW sensor stage and is further affected by the ISP and nonlinear camera response function (CRF).
Third, synthetic methods struggle to model realistic sensor noise and saturated pixel caused by the limited sensor dynamic range, both of which commonly appear in real scenes. 
\subsubsection{Acquisition Flexibility.} 
The primary challenge in capturing real-world blurry--sharp pairs is ensuring spatial and photometric alignment. 
Existing methods handle it through carefully designed hardware systems.
RBVD~\cite{zhu2022deep} uses a robotic arm to precisely control camera position and motion, capturing aligned short- and long-exposure pairs.
But its large robotic platform and limited mobility restrict scene diversity and hinder the scalability of data collection. 
Other methods~\cite{rim2020real,zhong2020efficient,rim2022realistic} employ beam splitters to synchronously capture short- and long-exposure pairs with two sensors.
But these systems require optical modifications, precise mechanical assembly, and careful calibration to ensure accurate alignment, making them bulky and less portable.
Moreover, differences in sensor response require additional post-processing to align the brightness and color of the captured pairs.

In this work, we aim to design a framework that achieves both blur realism and acquisition flexibility. 
Specifically, the framework should (1) capture real-world blurry images, (2) the capture device should be portable, (3) the data pairs should be aligned while the sharp images should be high-quality. 
To this end, we propose GS-RealBlur, which achieves the above expectations by building a bridge between captured 2D videos and real 3D scenes, as shown in Figure~\ref{fig:pipeline}. The acquisition pipeline is described below.

\subsection{Data Acquisition Pipeline}
\label{sec:3.2}
\subsubsection{Capturing Data.} 
We capture videos using an iPhone 16 Pro Max in manual mode. We record each scene in both handheld and gimbal-mounted configurations: the handheld setup naturally introduces motion blur, whereas the gimbal setup keeps the frames sharp. For sharp videos, we move the camera slowly and smoothly to maintain substantial overlap between adjacent frames. All videos are captured at 4K resolution. These dense, high-resolution sharp frames facilitate high-quality 3D reconstruction. Besides, the two setups share the same imaging parameters (\eg, ISO, aperture, shutter speed, and white balance) to maintain the photometric consistency across video frames and blurry-sharp images.
\subsubsection{Aligned Sharp Image Rendering.} 
Although the captured sharp frames may be not aligned with the blurry frames, they are sufficiently dense to cover nearly the entire scene content. Therefore, these sharp frames contain the ground truth information required to recover the blurry frames, and we extract it by reconstructing a 3D representation. After reconstructing the 3D representation, we can render a sharp image aligned with the blurry image. Specifically, we first use GLOMAP~\cite{pan2024global} to calibrate the camera pose of each blurry image within the 3D. Then, we refine the pose by the Blur-aware Pose Refinement (BPR) module to obtain more accurate one. With the refined pose, we render a sharp image from the reconstructed 3D representation and use it as supervision for training deblurring models.
\subsubsection{Artifact Filtering.} 
Despite our careful 3D reconstruction, novel-view renderings may still contain artifacts.
To ensure reliable supervision, we adopt a three-stage filtering strategy.
First, at the scene level, we compute the average PSNR between the ground-truth images and the corresponding rendered sharp images at held-out views. Then, we discard scenes whose average PSNR is below 36,dB, following GS-Blur~\cite{lee2024gs}.
Second, at the image level, we assess the quality of each sharp image rendered from the retained scenes using MUSIQ~\cite{ke2021musiq}. 
Images with a MUSIQ score below 63 are excluded.
Third, we manually inspect the remaining images to remove samples containing noticeable artifacts that are not detected by automatic filtering, further guaranteeing the overall dataset quality.

\subsection{Blur-aware Pose Refinement (BPR)}
\label{sec:3.3}
Existing structure-from-motion methods~\cite{schonberger2016structure, pan2024global} estimate camera poses by matching local keypoints, relying on high-frequency textures. However, for blurry images, blur degradation smooths image textures, reducing matching reliability and leading to inaccurate pose estimation. Therefore, we propose a Blur-aware Pose Refinement (BPR) module to optimize the poses estimated by GLOMAP~\cite{pan2024global}. 
\subsubsection{Overview of BPR.} 
First, for each blurry image $\mathbf B$, we estimate an initial pose $\mathbf P_0=(\mathbf R_0, \mathbf T_0)$ with GLOMAP~\cite{pan2024global}. The optimized pose $\mathbf P =(\mathbf R, \mathbf T)$ can be formulated as:
\begin{equation} 
    \mathbf{R} = \rm{\Delta} \mathbf{R} \cdot \mathbf R_0, \mathbf T = \mathbf T_0 +  \rm{\Delta} \mathbf{T},
\end{equation}
where $ \rm{\Delta} \mathbf{R}$ and $ \rm{\Delta} \mathbf{T}$ denote the learnable rotation increment and translation increment, and $\cdot$ denotes the matrix multiplication. Subsequently, given the optimized pose $\mathbf P$, we render a sharp image $\mathbf S$ from the reconstructed 3D representation with frozen parameters. 
Finally, the blurry image $\mathbf B$ and the sharp image $\mathbf S$ are used to compute the appearance consistency loss and the centroid-alignment regularizer to iteratively update $\rm{\Delta} \mathbf{R}$ and $\rm{\Delta} \mathbf{T}$. 
\subsubsection{Appearance Consistency Loss.} 
Although the rendered sharp image exhibits richer high-frequency details than the blurry one, they should have highly consistent low-frequency structures. Therefore, the appearance consistency loss is formulated as:
\begin{equation}
\mathcal{L}_{appearance}=\left\|\mathbf{S}_{\downarrow}-\mathbf{B}_{\downarrow}\right\|_{1},
\end{equation}
where $\downarrow$ denotes a $\times4$ downsampling operator, $\mathbf{S}$ denotes the rendered sharp image, and $\mathbf{B}$ denotes the blurry image. 
\subsubsection{Centroid Alignment Regularization.}
In our experiments, we find that although the appearance consistency loss provides effective content-aware supervision, it remains insufficient for translation alignment. We therefore introduce a centroid alignment regularizer to provide an additional geometric constraint.
Specifically, we first estimate the blur kernel between the rendered sharp image and the blurry observation by solving~\cite{cho2017convergence,xu2013unnatural,cho2009fast},:
\begin{equation}
    \label{eq:blur-kernel}
    \hat{\mathbf{K}} = \arg\min_{\mathbf{K}}  \|\mathbf{K} * \nabla \mathbf{S}-\nabla \mathbf{B}\|_2^2+\lambda\|\nabla \mathbf{K}\|_2^2,
\end{equation}
where $\mathbf{K}$ denotes the blur kernel, 
$*$ denotes convolution operation, $\nabla$ denotes a gradient operator, and $\lambda=10^3$ denotes a regularization weight. 
Then, we calculate the centroid $c$ and geometric center $c_0$ of the blur kernel $\mathbf{K}$:
\begin{equation}
    c=
    \left(
    \frac{\sum_{i,j} i\,\hat{\mathbf{K}}(i,j)}{\sum_{i,j} \hat{\mathbf{K}}(i,j)},
    \frac{\sum_{i,j} j\,\hat{\mathbf{K}}(i,j)}{\sum_{i,j} \hat{\mathbf{K}}(i,j)}
    \right), \ \ \
    c_0=\left(\frac{H}{2}, \frac{W}{2}\right),
\end{equation}
where $(i, j)$ denotes the spatial coordinates of the kernel elements. $H$ and $W$ represent the height and width of the blur kernel, respectively. Finally, centroid alignment loss is defined as:
\begin{equation}
    \mathcal{L}_{centroid}=\|c-c_0\|_2.
\end{equation}
\subsubsection{Final Learning Objective.} We optimize the camera pose by jointly minimizing the appearance consistency loss and the centroid alignment regularization:
\begin{equation}
    \mathcal{L}=(1-\alpha) \mathcal{L}_{appearance} + \alpha \mathcal{L}_{\mathrm{centroid}}.
\end{equation}
where $\alpha$ is a weighting coefficient set to 0.1.

\begin{table*}[t!] 
\centering
\caption{\textbf{Quantitative Comparison of Cross-Validation on Real-world Datasets.} Models are separately trained on different training sets and evaluate across various real-world benchmarks. Best results are in \textcolor{red}{red} and second-best are in \textcolor{blue}{blue}. Except for cases where training and testing sets match, the model trained on our GS-RealBlur achieves the best performance across all benchmarks.}
\label{tab:cross_dataset}
\scriptsize
\setlength{\tabcolsep}{2.5pt}
\resizebox{0.9\textwidth}{!}{%
\begin{tabular}{cc|ccc|ccc|ccc|ccc|ccc}
\toprule
\multicolumn{2}{c|}{%
\multirow{2}{*}{%
\diagbox{\textbf{Train set}}{\textbf{Test set}}}}
& \multicolumn{3}{c|}{RealBlur}
& \multicolumn{3}{c|}{RBVD}
& \multicolumn{3}{c|}{RSBlur} 
& \multicolumn{3}{c|}{BSD}
& \multicolumn{3}{c}{Average}  \\
& & PSNR & SSIM & LPIPS
& PSNR & SSIM & LPIPS
& PSNR & SSIM & LPIPS
& PSNR & SSIM & LPIPS
& PSNR & SSIM & LPIPS    \\
\midrule

\multirow{3}{*}{RealBlur}
& NAFNet
& \textcolor{red}{28.89} & \textcolor{red}{0.907} & 0.151 
& 26.00 & 0.892 & 0.249 
& 30.61 & 0.824 & 0.342 
& 30.00 & 0.914 & 0.125
& 28.88 & 0.884 & 0.221  \\
& Restormer
& \textcolor{red}{29.27} & \textcolor{red}{0.878} & \textcolor{red}{0.257}
& 26.08 & 0.821 & 0.313
& 29.85 & 0.778 & 0.429
& 30.31 & 0.903 & 0.211
& 28.88 & 0.845 & 0.303
\\
& EVSSM
& \textcolor{red}{29.59} & \textcolor{red}{0.918} & \textcolor{red}{0.239}
& 25.90 & 0.892 & 0.301
& 30.23 & 0.829 & 0.440
& 30.13 & 0.914 & 0.227
& 28.96 & 0.888 & 0.302
\\
\midrule
\multirow{3}{*}{RBVD}
& NAFNet
& 27.16 & 0.863 & 0.228
& \textcolor{blue}{26.51} & \textcolor{blue}{0.907} & 0.231 
& 29.59 & 0.793 & 0.388 
& 29.36 & 0.907 & 0.136
& 28.16 & 0.867 & 0.246
\\
& Restormer
& 26.93 & 0.814 & 0.333
& \textcolor{blue}{26.66} & \textcolor{blue}{0.835} & \textcolor{blue}{0.289}
& 29.08 & 0.747 & 0.417
& 29.17 & 0.892 & 0.220
& 27.96 & 0.822 & 0.315
\\
& EVSSM
& 26.98 & 0.871 & 0.329
& \textcolor{blue}{26.62} & \textcolor{blue}{0.909} & \textcolor{blue}{0.283}
& 29.71 & 0.814 & 0.442
& 28.73 & 0.893 & 0.236
& 28.01 & 0.872 & 0.323
\\
\midrule
\multirow{3}{*}{RSBlur}
& NAFNet
& 27.23 & 0.871 & 0.180  
& 26.36 & 0.905 & 0.234
& \textcolor{red}{33.72} & \textcolor{red}{0.877} & \textcolor{red}{0.310}
& 30.78 & 0.923 & 0.119
& \textcolor{blue}{29.54} & 0.894 & 0.211
\\
& Restormer
& 27.30 & 0.829 & 0.297
& 26.50 & 0.834 & 0.296
& \textcolor{red}{33.42} & \textcolor{red}{0.833} & \textcolor{red}{0.381}
& 30.81 & 0.910 & 0.211
& 29.51 & 0.852 & 0.296
\\
& EVSSM
& 27.44 & 0.876 & 0.287
& 26.42 & 0.904 & 0.300
& \textcolor{red}{34.02} & \textcolor{red}{0.875} & \textcolor{red}{0.354}
& 31.47 & 0.934 & 0.204
& \textcolor{blue}{29.84} & \textcolor{blue}{0.897} & 0.286
\\
\midrule
\multirow{3}{*}{BSD}
& NAFNet
& 26.88 & 0.864 & 0.217 
& 26.36 & 0.902 & 0.248
& 30.93 & 0.832 & 0.377 
& \textcolor{red}{33.87} & \textcolor{red}{0.952} & \textcolor{red}{0.078} 
& 29.53 & 0.888 & 0.230  
\\
& Restormer
& 27.01 & 0.823 & 0.316
& 26.44 & 0.825 & 0.318
& 31.19 & 0.792 & 0.421
& \textcolor{red}{33.68} & \textcolor{red}{0.943} & \textcolor{red}{0.167}
& \textcolor{blue}{29.58} & 0.846 & 0.306
\\
& EVSSM
& 25.87 & 0.847 & 0.328
& 26.36 & 0.900 & 0.284
& 30.32 & 0.829 & 0.425
& \textcolor{red}{36.10} & \textcolor{red}{0.964} & \textcolor{red}{0.138}
& 29.66 & 0.885 & 0.294
\\
\midrule
\multirow{3}{*}{GS-Blur}
& NAFNet
& 27.33 & 0.879 & \textcolor{blue}{0.147} 
& 26.26 & 0.904 & \textcolor{blue}{0.201}
& 32.87 & 0.860 & 0.317 
& 31.37 & 0.934 & 0.109
& 29.46 & \textcolor{blue}{0.895} & \textcolor{blue}{0.192} 
\\
& Restormer
& 27.32 & 0.841 & 0.276
& 26.52 & 0.833 & 0.293
& 32.42 & 0.817 & 0.403
& 31.26 & 0.920 & 0.205
& 29.38 & \textcolor{blue}{0.853} & \textcolor{blue}{0.294}
\\
& EVSSM
& 27.37 & 0.880 & 0.250
& 26.44 & 0.905 & 0.290
& 33.28 & 0.864 & 0.383
& 31.80 & 0.938 & 0.196
& 29.72 & \textcolor{blue}{0.897} & \textcolor{blue}{0.280}
\\
\midrule
\multirow{3}{*}{\textbf{GS-RealBlur} }
& NAFNet
& \textcolor{blue}{27.67} & \textcolor{blue}{0.886} & \textcolor{red}{0.140}
& \textcolor{red}{26.71} & \textcolor{red}{0.910} & \textcolor{red}{0.186}
& \textcolor{blue}{33.15} & \textcolor{blue}{0.863} & \textcolor{blue}{0.311}
& \textcolor{blue}{31.92} & \textcolor{blue}{0.939} & \textcolor{blue}{0.093}
& \textcolor{red}{29.86} & \textcolor{red}{0.900} & \textcolor{red}{0.183}
\\
& Restormer
& \textcolor{blue}{27.65} & \textcolor{blue}{0.852} & \textcolor{blue}{0.271}
& \textcolor{red}{26.85} & \textcolor{red}{0.838} & \textcolor{red}{0.280}
& \textcolor{blue}{32.59} & \textcolor{blue}{0.820} & \textcolor{blue}{0.400}
& \textcolor{blue}{31.63} & \textcolor{blue}{0.925} & \textcolor{blue}{0.203}
& \textcolor{red}{29.68} & \textcolor{red}{0.859} & \textcolor{red}{0.288}
\\
& EVSSM
& \textcolor{blue}{27.82} & \textcolor{blue}{0.887} & \textcolor{blue}{0.243}
& \textcolor{red}{26.96} & \textcolor{red}{0.911} & \textcolor{red}{0.277}
& \textcolor{blue}{33.48} & \textcolor{blue}{0.866} & \textcolor{blue}{0.380}
& \textcolor{blue}{32.17} & \textcolor{blue}{0.942} & \textcolor{blue}{0.193}
& \textcolor{red}{30.11} & \textcolor{red}{0.902} & \textcolor{red}{0.273}
\\
\bottomrule
\end{tabular}%
}
\vspace{-2mm}
\end{table*}

\subsection{Data Augmentation for Object Motion Blur}
\label{sec:3.4}

The primary limitation of using 3DGS is its restriction to rendering static scenes, meaning our method mainly addresses motion blur caused by camera movement.
However, in real-world blurry images, object motion blur and camera motion blur are independent, specifically manifested as inconsistent local blurring.
To alleviate the scarcity of object motion blur in our dataset, we suggest CutMix-Dynamic, a training-time data augmentation strategy inspired by CutMix~\cite{yun2019cutmix}.
Unlike the original CutMix, which randomly samples image patches, CutMix-Dynamic focuses on dynamic objects to better reflect real-world physical phenomena.
Specifically, we first use YOLOv8~\cite{varghese2024yolov8} to extract spatially aligned blurry-sharp patch pairs containing dynamic objects. 
These patches preserve characteristic dynamic textures and real motion blur patterns, and are collected into a dynamic object bank for later sampling. 
Then, during training, we sample a pair from the bank and paste the blurry and sharp patches onto the blurry input and its sharp supervision, to simulate the inconsistent local blurring.  
Consequently, training with CutMix-Dynamic improves the generalization of the deblurring model to real-world blurry images.
More implementation details and discussion are provided in the \textit{Suppl}.

\section{Experiments}
\subsection{Implementation Details}
\label{sec:4.1}
\subsubsection{Datasets.} 
To demonstrate the effectiveness of our proposed data acquisition framework, we train a representative deblurring model on different real-world datasets in a supervised manner respectively and evaluate its performance across multiple real-world benchmarks.
Specifically, six datasets with ground truth references (\ie, RealBlur~\cite{rim2020real}, RSBlur~\cite{rim2022realistic}, RBVD~\cite{zhu2022deep}, BSD~\cite{zhong2020efficient}, GS-Blur~\cite{lee2024gs}, and our dataset constructed in GS-RealBlur) are utilized for cross-dataset evaluation. 
Furthermore, we assess the out-of-distribution (OOD) generalization ability on two datasets without ground truth (\ie, RWBI~\cite{zhang2020deblurring} and DVD-Test~\cite{su2017deep}). 

\subsubsection{Training Details.} 
We adopt NAFNet~\cite{chen2022simple} as the deblurring network, which is trained with the AdamW optimizer~\cite{loshchilov2017decoupled} with $\beta_1=0.9$ and $\beta_2=0.999$ for 200k iterations.
Cosine annealing strategy~\cite{loshchilov2017sgdr} is employed to steadily decrease the learning rate from $10^{-3}$ to $10^{-6}$.
We randomly crop patches and augment them with flips and rotations.
The patch size is set to $256 \times 256$ and the batch size is set to $8$.
All experiments are conducted with PyTorch~\cite{paszke2019pytorch} on an Nvidia GeForce RTX A6000 GPU.

\subsubsection{Evaluation Configurations.}
For cross-dataset evaluation (\ie, RealBlur~\cite{rim2020real}, RBVD~\cite{zhu2022deep}, RSBlur~\cite{rim2022realistic} and BSD~\cite{zhong2020efficient}), we use PSNR, SSIM~\cite{wang2004image}, and LPIPS~\cite{zhang2018unreasonable} as the evaluation metrics.
For the evaluation of OOD generalization ability, we employ four recent no-reference metrics (\ie, MUSIQ~\cite{ke2021musiq}, MANIQA~\cite{yang2022maniqa} and CLIP-IQA~\cite{wang2023exploring}, as there is no ground truth in the benchmarks (\ie, RWBI~\cite{zhang2020deblurring} and DVD-Test~\cite{su2017deep}).

\subsection{Generalization on Real-World Scenarios}
\label{sec:4.2}

\begin{figure}[t!]
    \centering
    \includegraphics[width=1\linewidth]{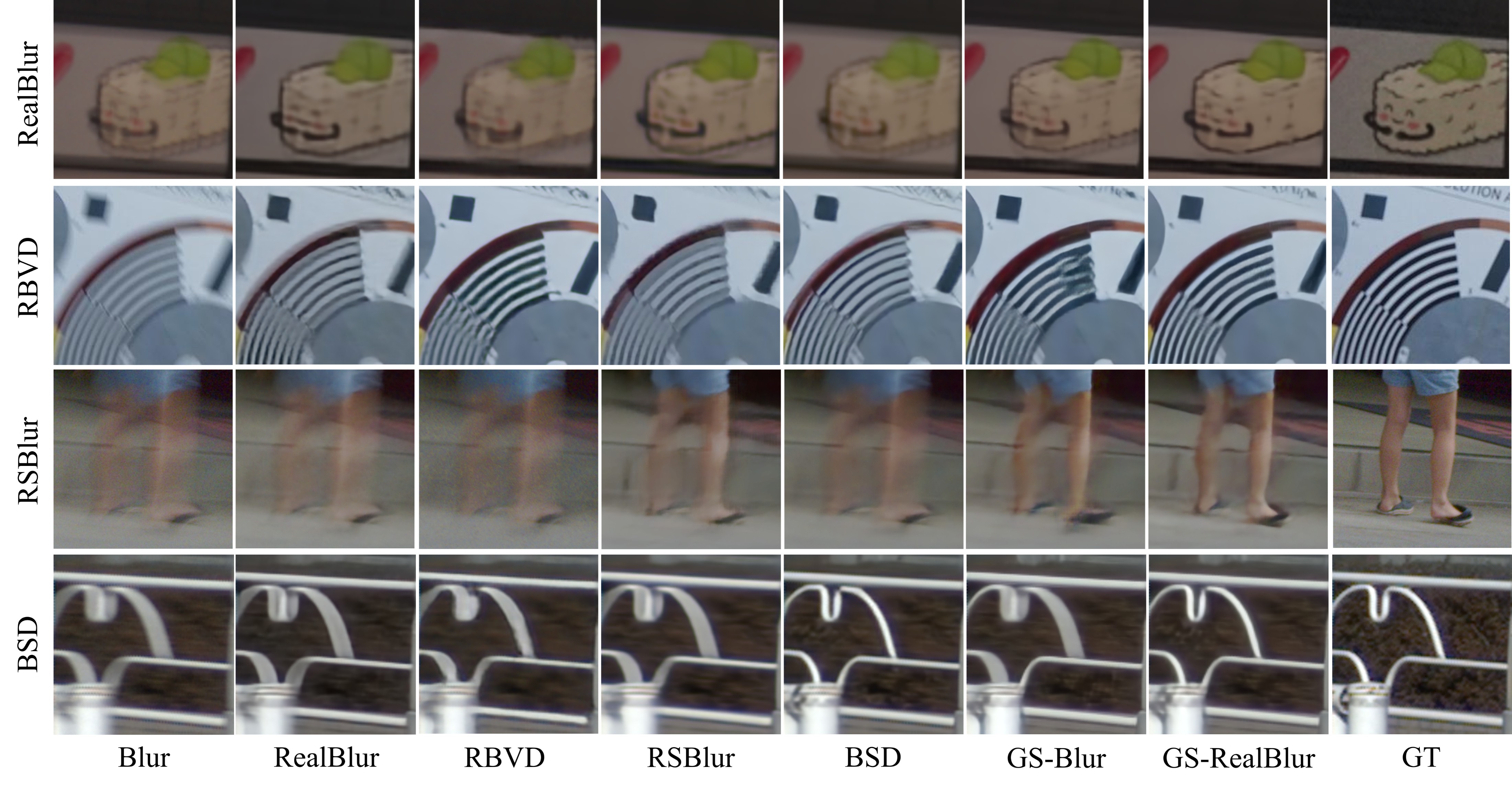}
    \caption{\textbf{Visual Comparison of Cross-Validation on Real-world Datasets}}
    \label{fig:metric}
    \vspace{-4mm}
\end{figure}

\subsubsection{Cross-Validation Results on Real-world Benchmarks.}
Table~\ref{tab:cross_dataset} reports the quantitative comparison of different training sets evaluated across various real-world benchmarks (\ie, RealBlur~\cite{rim2020real}, RBVD~\cite{zhu2022deep}, RSBlur~\cite{rim2022realistic} and BSD~\cite{zhong2020efficient}).
The results indicate a severe domain gap among existing datasets, as models trained on themselves suffer significant performance drops when tested on others. 
In contrast, our GS-RealBlur remarkably alleviates this issue. 
Even without training on the target domain, the model trained on GS-RealBlur delivers highly competitive cross-domain performance, securing the highest average scores among all training sets. 
This strong performance highlights the superior quality and diversity of our dataset constructed via the proposed framework.
Visual comparisons in Figure~\ref{fig:metric} further support our claims.
In cross-domain scenarios (where training and testing sets differ), the model trained on GS-RealBlur consistently produces the most visually pleasing results. 
Remarkably, even in in-domain settings, it exhibits visual quality equivalent to or even higher than the models specifically optimized on those target datasets.

\subsubsection{Generalization Results on In-the-wild Data.}
\begin{table}[t!]
\centering
\caption{\textbf{Quantitative Comparison of In-the-wild Data}}
\label{tab:nr-metric}
\resizebox{\linewidth}{!}{
\begin{tabular}{c|ccc|ccc}
\toprule
\multirow{2}{*}{\diagbox{\textbf{Train set}}{\textbf{Test set}}} 
& \multicolumn{3}{c|}{\textbf{RWBI}}
& \multicolumn{3}{c}{\textbf{DVD-Test}} \\
& MUSIQ$\uparrow$ & MANIQA$\uparrow$ & CLIP-IQA$\uparrow$ & MUSIQ$\uparrow$ & MANIQA$\uparrow$ & CLIP-IQA$\uparrow$ \\
\midrule
RealBlur & 58.552 & 0.266 & 0.341  & 45.040 & 0.231 & 0.290  \\
RBVD & 52.014 & 0.258 & 0.313  & 40.970 & 0.206 & 0.224  \\
RSBlur & 57.929 & 0.264 & 0.335  & 41.132 & 0.217 & 0.274  \\
BSD & 58.104 & 0.273 & 0.336  & 40.595 & 0.214 & 0.275  \\
GS-Blur & 61.330 & 0.295 & 0.367  & 45.371 & 0.235 & 0.284  \\
\midrule
\textbf{GS-RealBlur} & \textbf{61.610} & \textbf{0.300} & \textbf{0.372} & \textbf{46.604} & \textbf{0.242} & \textbf{0.294} \\
\bottomrule
\end{tabular}}
\vspace{-4mm}
\end{table}
To further assess the out-of-distribution (OOD) generalization on in-the-wild data, we conduct evaluations on the RWBI~\cite{zhang2020deblurring} and DVD-Test~\cite{su2017deep} datasets in NAFNet~\cite{chen2022simple}. 
As reported in Table~\ref{tab:nr-metric}, the model trained on our GS-RealBlur demonstrates remarkable robustness, consistently achieving the highest scores across all metrics on both test sets and outperforming models trained on existing benchmarks by a clear margin. 

\subsection{Ablation Studies}
\label{sec:4.3}
We conduct comprehensive ablation studies to validate the effectiveness of each component in the proposed framework. 
Unless otherwise specified, all ablation experiments are trained on 5,000 samples using NAFNet~\cite{chen2022simple}, following the same training setting described in Training Details. 
Due to space limitations, the complete experimental results are provided in the supplementary material, while only selected metrics are reported here.

\subsubsection{Effect of BPR and Artifact Filtering.}
The effect of Blur-aware Pose Refinement (BPR) and artifact filtering are summarized in Table~\ref{tab:component_ablation}. 
Without both BPR and artifact filtering, the constructed pairs suffer from geometric misalignment and rendering artifacts, leading to poor deblurring performance.
BPR improves supervision alignment leading to performance improvement, indicating the importance of accurate pose estimation for supervision construction.
Artifact filtering removes samples with severe artifacts, also enabling performance gain.
Their combination performs best.

\begin{table}[t!]
\centering
\caption{\textbf{Ablation on BPR and Artifact Filtering}}
\vspace{-2mm}
\label{tab:component_ablation}
\resizebox{\linewidth}{!}{
\begin{tabular}{c c|cc|cc|cc|cc}
\toprule
\multirow{2}{*}{BPR} &  \multirow{2}{*}{\makecell[c]{Artifact \\ Filtering}}
& \multicolumn{2}{c|}{RealBlur}
& \multicolumn{2}{c|}{RBVD}
& \multicolumn{2}{c|}{RSBlur}
& \multicolumn{2}{c|}{BSD} \\
& & PSNR$\uparrow$ & SSIM$\uparrow$ & PSNR$\uparrow$ & SSIM$\uparrow$ & PSNR$\uparrow$ & SSIM$\uparrow$ & PSNR$\uparrow$ & SSIM$\uparrow$ \\
\midrule
\textcolor{darkred}{\usym{2717}} & \textcolor{darkred}{\usym{2717}} & 26.36 & 0.848 & 26.06 & 0.892 & 30.89 & 0.821 & 29.92 & 0.917 \\
\textcolor{darkgreen}{\usym{2713}} & \textcolor{darkred}{\usym{2717}} & 27.13 & 0.863 & 26.33 & 0.897 & 32.41 & 0.853 & 30.86 & 0.928  \\
\textcolor{darkred}{\usym{2717}} & \textcolor{darkgreen}{\usym{2713}} & 27.38 & 0.880 & 26.46 & 0.899 & 32.61 & 0.854 & 31.42 & 0.930 \\
\textcolor{darkgreen}{\usym{2713}} & \textcolor{darkgreen}{\usym{2713}} & \textbf{27.57} & \textbf{0.885} & \textbf{26.67} & \textbf{0.908} & \textbf{33.13} & \textbf{0.861} & \textbf{31.63} & \textbf{0.936} \\
\bottomrule
\end{tabular}
}
\vspace{-2mm}
\end{table}

\subsubsection{Effect of Loss Terms in BPR.}
\begin{figure}[t!]
    \centering
    \includegraphics[width=1\linewidth]{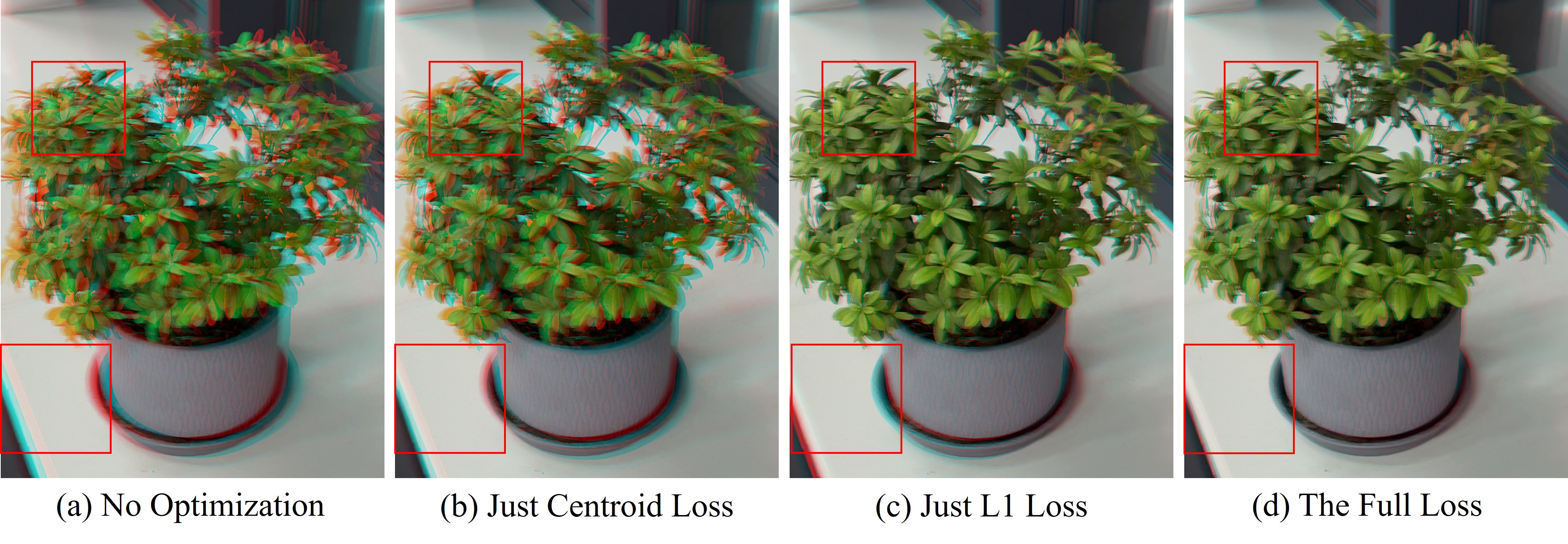}
    \caption{\textbf{Effect of BPR on Geometric Alignment.}  Each visualization result is shown as a stereo-anaglyph image.
    `Full loss' produces more aligned results around the plant leaves, table edges, and the tray.} 
    \label{fig:BPR-compare}
    \vspace{-3mm}
\end{figure}

\begin{table}[t]
\centering
\caption{\textbf{Ablation on Loss Terms in BPR.}}
\vspace{-2mm}
\label{tab:pose_loss_ablation}
\resizebox{\linewidth}{!}{
\begin{tabular}{c c|c|cc|cc|cc|cc}
\toprule
\multirow{2}{*}{$\mathcal{L}_{appearance}$} & \multirow{2}{*}{$\mathcal{L}_{\text{centroid}}$ } 
& \multicolumn{1}{c|}{Pose Error}
& \multicolumn{2}{c|}{RealBlur}
& \multicolumn{2}{c|}{RBVD}
& \multicolumn{2}{c|}{RSBlur}
& \multicolumn{2}{c}{BSD} \\
& & TE$\downarrow$ / RE$\downarrow$ 
& PSNR$\uparrow$ & SSIM$\uparrow$ 
& PSNR$\uparrow$ & SSIM$\uparrow$ 
& PSNR$\uparrow$ & SSIM$\uparrow$ 
& PSNR$\uparrow$ & SSIM$\uparrow$  \\
\midrule
\textcolor{darkred}{\usym{2717}} & \textcolor{darkred}{\usym{2717}} 
& 1.036 / 1.102 
& 27.38 & 0.880 & 26.46 & 0.899 & 32.61 & 0.854 & 31.42 & 0.930 \\
\textcolor{darkgreen}{\usym{2713}} & \textcolor{darkred}{\usym{2717}}
& 0.702 / 0.503
& 27.52 & 0.884 & 26.63 & 0.907 & 32.95 & 0.858 & 31.59 & 0.935 \\
\textcolor{darkred}{\usym{2717}} & \textcolor{darkgreen}{\usym{2713}}
& 0.452 / 0.841
& 27.43 & 0.883 & 26.52 & 0.903 & 32.75 & 0.856 & 31.51 & 0.934 \\
\textcolor{darkgreen}{\usym{2713}} & \textcolor{darkgreen}{\usym{2713}} 
& \textbf{0.445} / \textbf{0.479}
& \textbf{27.57} & \textbf{0.885}  & \textbf{26.67} & \textbf{0.908} & \textbf{33.13} & \textbf{0.861} & \textbf{31.63} & \textbf{0.936} \\

\bottomrule
\end{tabular}
}
\vspace{-2mm}
\end{table}
We analyze the contribution of each loss term in BPR using pose errors and downstream deblurring performance. 
Since directly measuring the alignment of real blurry--sharp pairs is challenging, we construct a synthetic evaluation set using GS-Blur and regard the calibrated poses of the corresponding sharp images as ground truth. 
We report Translation Error (TE) and Rotation Error (RE) to evaluate alignment quality.
As shown in Table~\ref{tab:pose_loss_ablation}, $\mathcal{L}_{\mathrm{centroid}}$ primarily reduces translation error but offers limited improvement in deblurring when used alone, while $\mathcal{L}_{\mathrm{appearance}}$ effectively reduces the rotation error, resulting in substantially improved deblurring performance. 
Combining them achieves the lowest pose errors and the best deblurring performance, demonstrating their complementarity.
Figure~\ref{fig:BPR-compare} further confirms their complementarity, where the full loss produces the most accurate alignment by jointly correcting spatial displacement and texture misalignment.

\subsubsection{Effect of CutMix-Dynamic.}
\begin{table}[t!]
\centering
\caption{\textbf{Ablation on CutMix-Dynamic}}
\vspace{-2mm}
\label{tab:dynamic_ablation}
\resizebox{\linewidth}{!}{
\begin{tabular}{l|cc|cc|cc|cc}
\toprule
\multicolumn{1}{c|}{\multirow{2}{*}{Methods}}
& \multicolumn{2}{c|}{RealBlur}
& \multicolumn{2}{c|}{RBVD}
& \multicolumn{2}{c|}{RSBlur}
& \multicolumn{2}{c}{BSD} \\
& PSNR$\uparrow$ & SSIM$\uparrow$
& PSNR$\uparrow$ & SSIM$\uparrow$
& PSNR$\uparrow$ & SSIM$\uparrow$
& PSNR$\uparrow$ & SSIM$\uparrow$\\
\midrule
Baseline & 27.52 & 0.884 & 26.60 & 0.908 & 33.09 & \textbf{0.860} & 31.52 & 0.935 \\
+ CutMix & 27.50 & 0.884 & 26.62 & 0.908 & 33.10 & \textbf{0.860} & 31.59 & \textbf{0.936} \\
+ CutMix-Dynamic & \textbf{27.58} & \textbf{0.885} & \textbf{26.67} & \textbf{0.909} & \textbf{33.12} & \textbf{0.860} & \textbf{31.64} & \textbf{0.936} \\
\bottomrule
\end{tabular}
}
\vspace{-2mm}
\end{table}
We conduct an ablation study to evaluate the effectiveness of CutMix-Dynamic, as shown in Table~\ref{tab:dynamic_ablation}.  The baseline performs consistently on both static and dynamic benchmarks. 
Introducing CutMix~\cite{yun2019cutmix} improves performance on the dynamic datasets but slightly degrades results on the static RealBlur dataset, likely due to physically implausible local replacements.
In contrast, by inserting textures extracted from dynamic objects, our CutMix-Dynamic achieves the best results on both datasets.
It more accurately models real motion blur while preserving static scene fidelity.

\subsubsection{Effect of Dataset Scale.}
To investigate the effect of dataset scale, we conduct an ablation study by training the model with different proportions of the training dataset.
Specifically, we randomly sample 25\%, 50\%, and 100\% of the training data while keeping all other training settings unchanged.
As shown in Table~\ref{tab:datascale_ablation}, a consistent performance improvement as the amount of training data increases. 
When trained with only 25\% of the data, the model already achieves competitive results. Increasing the training data to 50\% leads to further improvements, and the model achieves the best performance when trained with the full dataset. 
This trend highlights the benefit of scaling up GS-RealBlur.

\begin{table}[t!]
\centering
\caption{\textbf{Ablation on Dataset Scale}}
\vspace{-2mm}
\label{tab:datascale_ablation}
\resizebox{\linewidth}{!}{
\begin{tabular}{c|cc|cc|cc|cc}
\toprule
\multirow{2}{*}{Data Proportions}
& \multicolumn{2}{c|}{RealBlur}
& \multicolumn{2}{c|}{RBVD}
& \multicolumn{2}{c|}{RSBlur}
& \multicolumn{2}{c}{BSD} \\
& PSNR$\uparrow$ & SSIM$\uparrow$ & PSNR$\uparrow$ & SSIM$\uparrow$ & PSNR$\uparrow$ & SSIM$\uparrow$  & PSNR$\uparrow$ & SSIM$\uparrow$ \\
\midrule
25\% & 27.52 & 0.883 & 26.47 & 0.907 & 33.08 & 0.862 & 31.58 & 0.935 \\
50\% & 27.58 & 0.885 & 26.57 & 0.908 & 33.10 & 0.862 & 31.69 & 0.937 \\
100\% & \textbf{27.67} & \textbf{0.886} & \textbf{26.71} & \textbf{0.910} &  \textbf{33.15} & \textbf{0.863} & \textbf{31.92} & \textbf{0.939} \\
\bottomrule
\end{tabular}
}
\vspace{-2mm}
\end{table}

\section{Conclusion}
Existing data acquisition methods face a trade-off between blur realism and acquisition flexibility.
With this motivation, we propose GS-RealBlur, a data acquisition framework that simultaneously achieves both objects for real-world image deblurring. 
GS-RealBlur first captures blurry and sharp images of a scene using consumer-grade devices in handheld and gimbal-mounted configurations.
Then, the sharp frames are used to reconstructed a 3D representation, from which sharp references are rendered for blurry images.
To address inaccurate pose under blur, we further introduce a Blur-aware Pose Refinement (BPR) module to optimize the pose by an appearance consistency loss and a centroid alignment regularizer. 
Finally, we use the data acquisition framework to construct a single-image deblurring dataset. 
Extensive experiments demonstrate that a deblurring model trained on our dataset exhibits superior generalization performance on real-world scenarios compared to models trained on existing datasets, validating the effectiveness of GS-RealBlur.

\clearpage
\appendix

\twocolumn[
\begin{center}
    \vspace{14mm}
    {\LARGE \textbf{GS-RealBlur: A Flexible Data Acquisition Framework for \\ Real-World Image Deblurring  (Supplementary Material)}}
    \vspace{13mm}
\end{center}
]

\renewcommand{\thesection}{\Alph{section}}
\renewcommand{\thetable}{\Alph{table}}
\renewcommand{\thefigure}{\Alph{figure}}
\renewcommand{\theequation}{\Alph{equation}}

\setcounter{section}{0}
\setcounter{figure}{0}
\setcounter{table}{0}
\setcounter{equation}{0}

\section{3D Gaussian Splatting}
\label{sec:3dgs}
3DGS models a scene with a set of three-dimensional Gaussian ellipsoids $\mathcal{K}$, where each Gaussian primitive $k \in \mathcal{K}$ is represented by a quadruple 
\begin{equation}
\{\boldsymbol{\mu}_k, \boldsymbol{\Sigma}_k, \sigma_k, \mathbf{S}_k\}_{k \in \mathcal{K}}
\end{equation}
where $\boldsymbol{\mu}_k \in \mathbb{R}^3$ denotes the spatial position, $\boldsymbol{\Sigma}_k \in \mathbb{R}^{3 \times 3}$ denotes the covariance matrix, $\sigma_k$ denotes the opacity parameter, and $\mathbf{S}_k$ denotes spherical harmonic coefficients that model view-dependent color variation. The initial Gaussian parameters are typically estimated using Structure-from-Motion (SfM)~\cite{schonberger2016structure,pan2024global} and are further refined in subsequent optimization.

During rendering, the three-dimensional Gaussian ellipsoids are projected onto the image plane. The color of each pixel $p$ is then computed using point-based $\alpha$-blending as follows:
\begin{equation}
\hat{C}(p) = \sum_{k \in \mathcal{K}} \alpha_k  \mathbf{c}(\mathbf{v}_k; \mathbf{S}_k) \prod_{j=1}^{k-1} (1 - \alpha_j),
\end{equation}
$\alpha_k$ denotes the opacity of the $k$-th Gaussian primitive. More detailed formulations of the imaging model are provided in~\cite{kerbl20233d}.

Compared with implicit volumetric rendering methods such as NeRF~\cite{mildenhall2021nerf}, 3DGS employs explicit Gaussian ellipsoids with efficient rasterization, enabling fast and high-quality rendering. In this work, we leverage its geometric consistency to generate sharp images from the reconstructed scene, providing stable and precisely aligned supervision for real blurry images.

\section{Discussion about Rendered Supervision} 
Unlike early works that use captured sharp images as supervision, our sharp references are rendered from reconstructed 3DGS representations.
This design enables accurate spatial alignment but also introduces concerns regarding the domain gap and the quality of rendered supervision.
We discuss the motivation, trade-off, and reliability of this design below.

\subsection{Motivation of the Design}
Obtaining accurately aligned blurry-sharp pairs in real-world scenes is challenging because the camera pose changes during exposure and may also vary between separate blurry and sharp captures.
Consequently, directly capturing aligned blurry and sharp pairs is difficult.
Post-processing methods like optical-flow alignment provide a possible solution, but they are often unreliable in heavily blurred regions and may introduce additional warping artifacts.
In contrast, 3DGS provides a high-fidelity scene representation and allows sharp images to be rendered from any viewpoints that are adaptively aligned with the blurry observations.
It therefore provides a practical way to preserve real captured blur while obtaining spatially aligned sharp supervision, which is difficult to achieve using conventional acquisition and alignment methods.

\subsection{Trade-Off of the Design}
This design relies on rendered sharp references. However, it preserves fully real captured blur and enables the construction of spatially aligned blurry-sharp pairs. 
It entails a clear trade-off between the authenticity of sharp supervision and the realism of input blur.
We argue that realistic blurry inputs are more important for downstream deblurring than strictly real-captured sharp supervision.
This argument is supported by two observations. 
First, as shown in Table 2 in the main paper, models trained on our dataset achieve the best generalization performance.
Second, we further construct two training sets from the same scenes. One uses blurry images synthesized by GS-Blur, whereas the other uses real blurry images captured by our framework.
In RealBlur~\cite{rim2020real} test set, the model trained with real blurry images achieves a PSNR/SSIM of 27.48/0.880, compared with 27.11/0.875 for the model trained with synthetic blurry images.
These results indicate that the benefit of realistic blurry inputs outweighs the rendering gap.

Besides, to ensure the reliability of rendered supervision as much as possible, we employ a carefully designed framework.
During data acquisition, we densely capture sharp-view videos to provide sufficient scene observations for 3DGS reconstruction. 
The blurry images are captured within the spatial range covered by sharp videos, reducing artifacts caused by large viewpoint deviations or insufficient observations.
We further crop the images into patches and remove samples with evident artifacts by metric and strict manual inspection, ensuring that only reliable references are retained for training.

Overall, our design preserves the real-captured blur while providing spatially aligned high-quality supervision.
Although the sharp supervisions are rendered, our experimental analyses show that the resulting trade-off is beneficial, supporting the reliability of rendering supervisions for deblurring.

\begin{figure*}[t]
    \centering
    \includegraphics[width=1\linewidth]{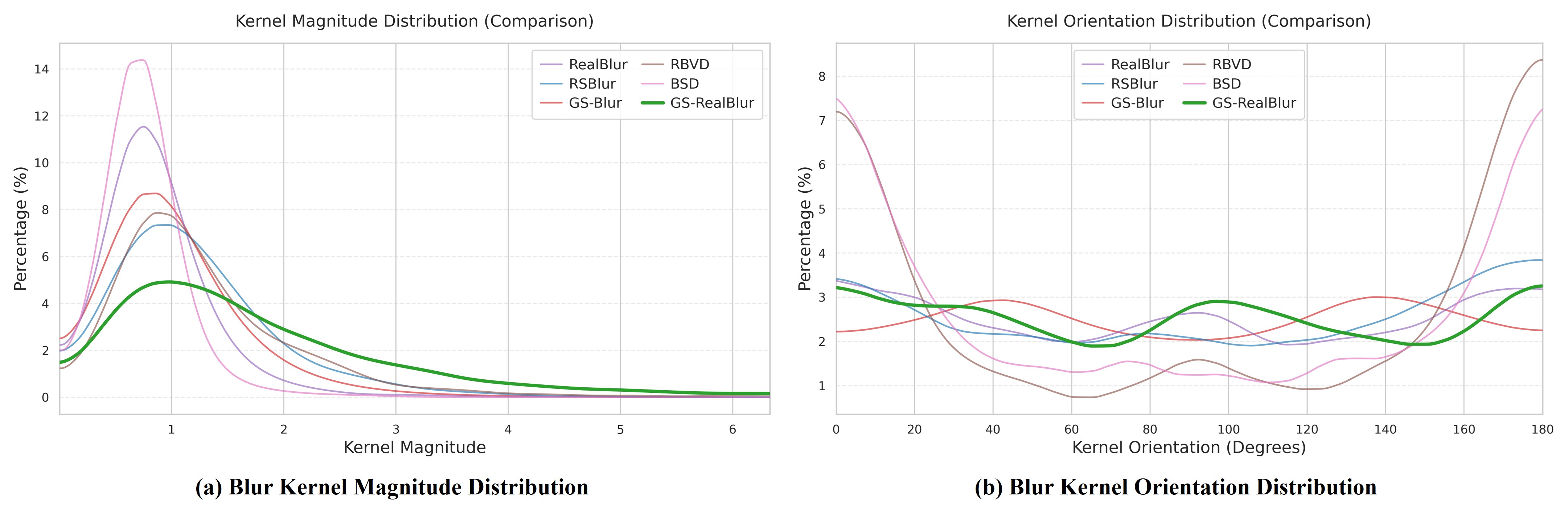}
    \caption{\textbf{Blur Distribution of Deblurring Datasets}} 
    \label{fig:blur-distribution}
\end{figure*}

\section{Blur Distribution of GS-RealBlur Dataset.} 
To characterize the diversity of blur patterns in GS-RealBlur, we estimate the blur kernel of each blurry image using Equation~(\ref{eq:blur-kernel-supply}) and analyze its magnitude and orientation.
\begin{equation}
    \label{eq:blur-kernel-supply}
    \hat{\mathbf{K}} = \arg\min_{\mathbf{K}}  \|\mathbf{K} * \nabla \mathbf{S}-\nabla \mathbf{B}\|_2^2+\lambda\|\nabla \mathbf{K}\|_2^2,
\end{equation}
where $\mathbf{K}$ donates the blur kernel, 
$*$ denotes convolution operation, $\nabla$ denotes a gradient operator, and $\lambda=10^3$ denotes a regularization weight. 

As shown in Figure~\ref{fig:blur-distribution}, benefiting from our flexible blurry images collection, GS-RealBlur exhibits a wider range of blur magnitudes and a more uniform distribution of blur orientations.

\section{Data Augmentation for Object Motion Blur}
\label{sec:CutMix-Dynamic}
\begin{figure*}[t]
    \centering
    \includegraphics[width=1\linewidth]{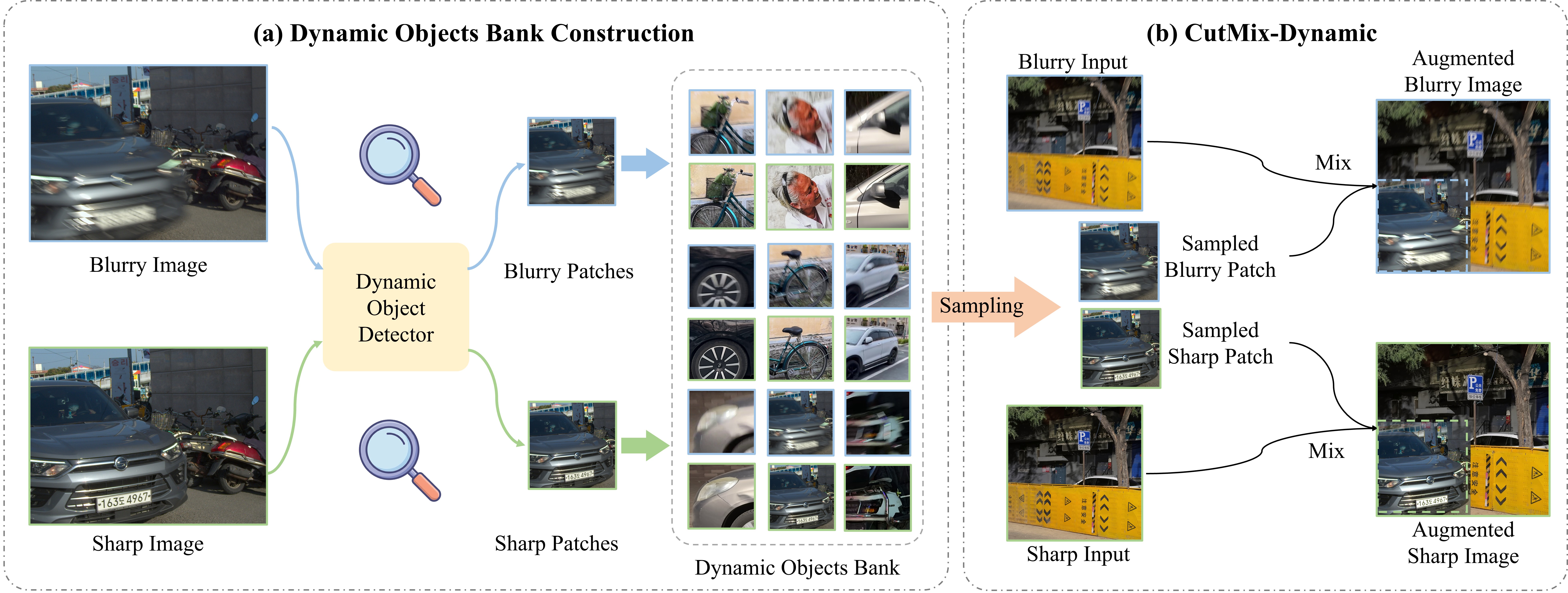}
    \caption{\textbf{Overview of our CutMix-Dynamic}. We first construct a dynamic objects bank by extracting blurry-sharp patch pairs containing dynamic objects using a dynamic object detector. Subsequently, during the training process, CutMix-Dynamic randomly samples a patch pair from the Dynamic Objects Bank and pastes the blurry and sharp patches on the input blurry image and its sharp supervision, respectively. } 
    \label{fig:cutmix-dynamic-structure}
\end{figure*}

The primary limitation of using 3DGS is its restriction to rendering static scenes, meaning our method can only model motion blur caused by camera movement. However, in real-world blurry images, object motion blur and camera motion blur are independent, specifically manifested as inconsistent local blurring.
To address the scarcity of dynamic blur in our datasets, we propose CutMix-Dynamic, a training-time data augmentation strategy inspired by CutMix~\cite{yun2019cutmix}. Unlike the original CutMit, which randomly extracts patches, CutMix-Dynamic specifically samples paired blurry and sharp patches containing dynamic objects. It therefore introduces realistic local motion blur while maintaining spatially aligned supervision. It could be done in 2 steps.

\subsubsection{Dynamic Object Bank Construction}
As illustrated in Figure~\ref{fig:cutmix-dynamic-structure}(a), first, we define 9 categories of dynamic objects~\cite{lin2014microsoft}, including person, bicycle, car, motorcycle, bus, truck, cat, dog and sports ball. Then, we use a dynamic object detector (\ie, YOLOv8~\cite{varghese2024yolov8}) to detect these objects and extract blurry-sharp patch pairs containing them from datasets. We collect these patch pairs to construct the Dynamic Objects Bank.
These patches contain characteristic dynamic textures and real motion blur patterns, and we collect them into a dynamic object bank.

\subsubsection{Cutmix-Dynamic}
As illustrated in Figure~\ref{fig:cutmix-dynamic-structure}(b), during training, we sample a patch pair from the bank, and paste the blurry patch on the input blurry images to simulate inconsistent local blurring. Simultaneously, the corresponding sharp patch is pasted on the ground truth sharp image to maintain consistent supervision. Consequently, training on the data enables the deblurring model to generalize effectively to real-world blurry images, even those with object motion blur.Specifically, we randomly sample a patch pair from the Dynamic Objects Bank and paste the blurry and sharp patches on the input blurry image and its sharp supervision, respectively. The augmented blurry images also exhibit locally inconsistent blur, similar to that observed in real-world images.

\section{More Experiments}
\subsection{Additional Visual Comparison}
\begin{figure*}[t]
    \centering
    \begin{subfigure}[t]{0.9\textwidth}
        \centering
        \includegraphics[width=\linewidth]{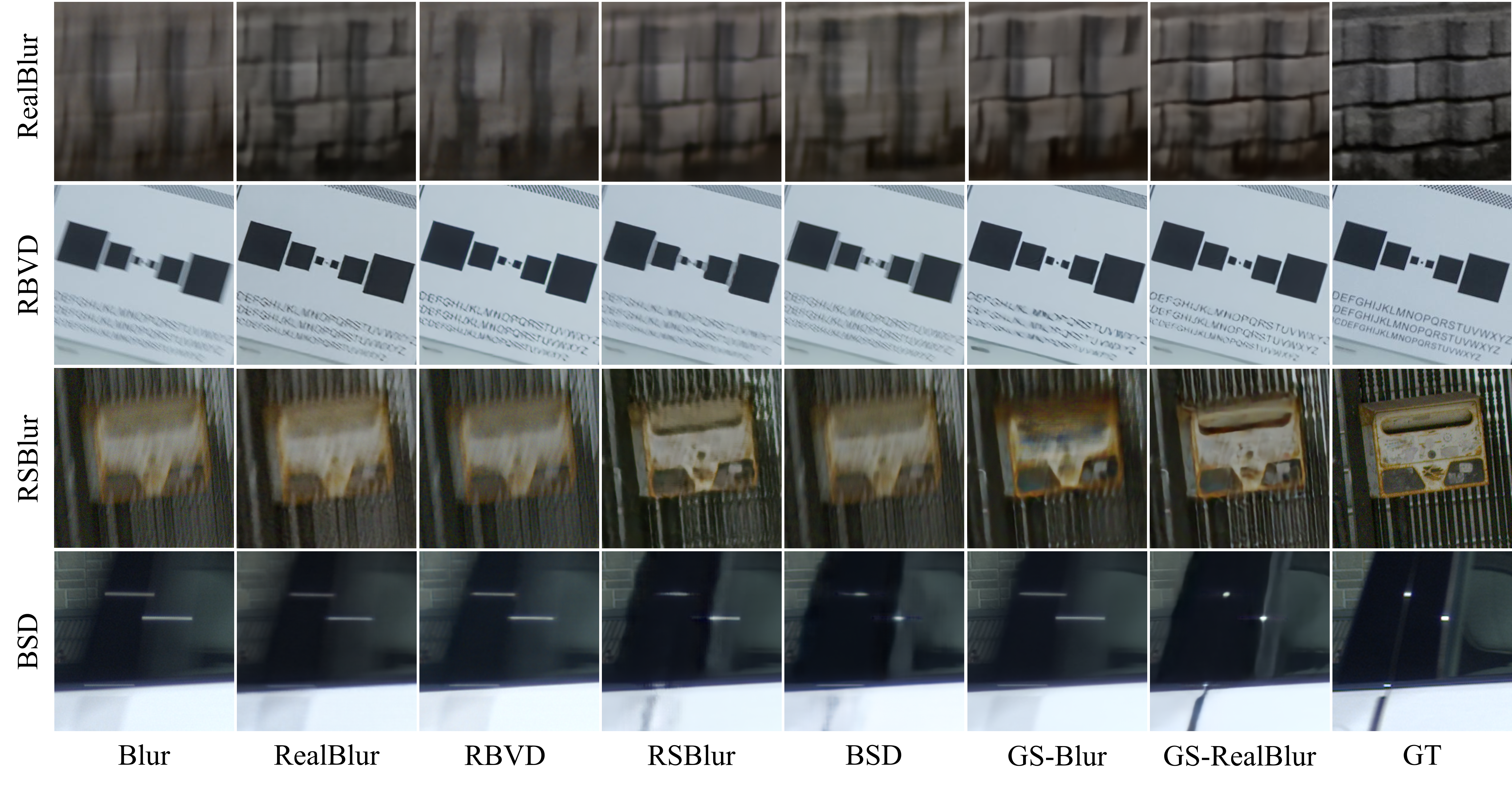}
        \caption{\textbf{Visual Comparison on Real-world Datasets}}
        \label{fig:more-metric-visual}
    \end{subfigure}

    \vspace{3mm}
    
    \begin{subfigure}[t]{0.9\textwidth}
        \centering
        \includegraphics[width=\linewidth]{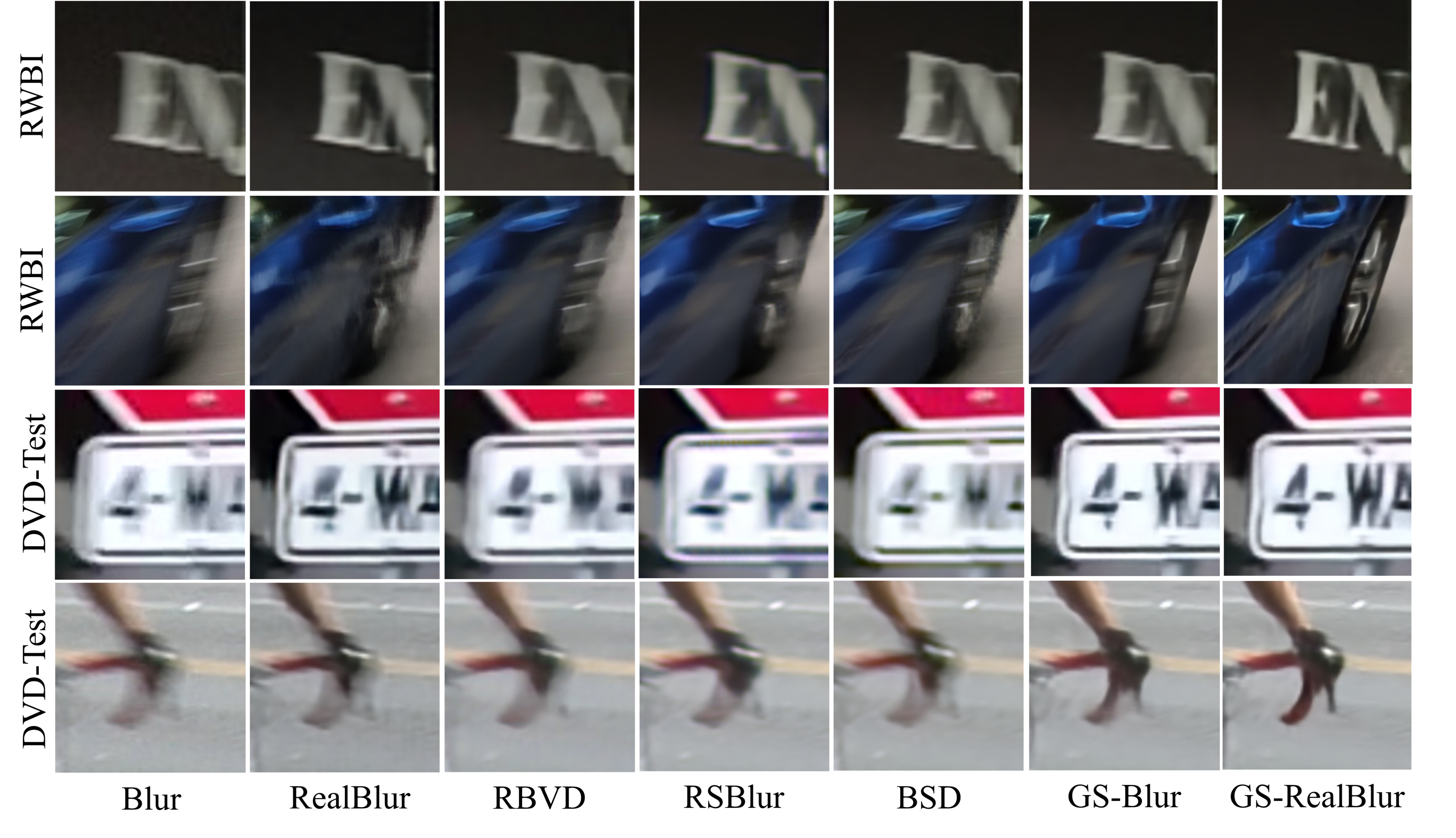}
        \caption{\textbf{Visual Comparison on In-the-wild Data}}
        \label{fig:nr-metric}
    \end{subfigure}

    \caption{\textbf{Additional visual comparisons.}}
    \label{fig:additional-visual-comparison}
    \vspace{-3mm}
\end{figure*}

Figure~\ref{fig:additional-visual-comparison} presents additional visual comparisons. Figure~\ref{fig:additional-visual-comparison} (a) presents comparisons on real-world datasets, and Figure~\ref{fig:additional-visual-comparison} (b) shows visual results in in-the-wild data. In (b), rows 1 and 3 contain camera-motion blur, while rows 2 and 4 contain object-motion blur. The model trained on GS-RealBlur produces the clearest results.

\subsection{Study of Artifact Filtering.}
To investigate the effect of different filtering thresholds and determine an appropriate MUSIQ threshold, we manually select 500 samples from 100 scenes for evaluation. 
Half of the samples are manually identified as high-quality, while the other half contain noticeable artifacts.
We then apply different MUSIQ thresholds to these samples and evaluate their ability to retain high-quality samples and reject samples with artifacts.
Thresholds of 60, 63, and 65 retain 80\%, 76\%, and 63\% of the high-quality patches, while rejecting 55\%, 61\%, and 64\% of the patches with artifacts, respectively. 
We therefore set the MUSIQ threshold to 63, as it achieves a reasonable trade-off between retaining high-quality samples and reducing the burden of manual artifact inspection.

\subsection{Full Ablation Studies Results}
\label{sec:ablation study}
Due to space limitations, we provide the complete ablation results in this section.
Unless otherwise specified, all ablation experiments are trained on 5,000 samples using NAFNet~\cite{chen2022simple}, following the same training setting described in Training Details in the main paper. 

\subsubsection{Effect of BPR and Artifact Filtering.}
Table~\ref{tab:component_ablation_sup} presents the complete results for BPR and artifact filtering.
Consistent with the observations in the main paper, each component independently improves deblurring performance by mitigating misalignment or excluding severely corrupted samples, while their combination achieves the best results.
\begin{table*}[t]
\centering
\caption{\textbf{Ablation on BPR and Artifact Filtering}}
\vspace{-2mm}
\label{tab:component_ablation_sup}
\resizebox{\linewidth}{!}{
\begin{tabular}{c c|ccc|ccc|ccc|ccc}
\toprule
\multirow{2}{*}{BPR} &  \multirow{2}{*}{\makecell[c]{Artifact \\ Filtering}}
& \multicolumn{3}{c|}{RealBlur}
& \multicolumn{3}{c}{RBVD}
& \multicolumn{3}{c}{RSBlur}
& \multicolumn{3}{c}{BSD} \\
& & PSNR$\uparrow$ & SSIM$\uparrow$ & LPIPS$\downarrow$ & PSNR$\uparrow$ & SSIM$\uparrow$ & LPIPS$\downarrow$ & PSNR$\uparrow$ & SSIM$\uparrow$ & LPIPS$\downarrow$ & PSNR$\uparrow$ & SSIM$\uparrow$ & LPIPS$\downarrow$\\
\midrule
\textcolor{darkred}{\usym{2717}} & \textcolor{darkred}{\usym{2717}} & 26.36 & 0.848 & 0.244 & 26.06 & 0.892 & 0.198 & 30.89 & 0.821 & 0.326 & 29.92 & 0.917 & 0.165 \\
\textcolor{darkgreen}{\usym{2713}} & \textcolor{darkred}{\usym{2717}} & 27.13 & 0.863 & 0.229 & 26.33 & 0.897 & 0.190 & 32.41 & 0.853 & 0.319 & 30.86 & 0.928 & 0.126 \\
\textcolor{darkred}{\usym{2717}} & \textcolor{darkgreen}{\usym{2713}} & 27.38 & 0.880 & 0.154 & 26.46 & 0.899  & 0.189 &  32.61 & 0.854 & 0.315 & 31.42 & 0.930 & 0.109 \\
\textcolor{darkgreen}{\usym{2713}} & \textcolor{darkgreen}{\usym{2713}} & \textbf{27.57} & \textbf{0.885} & \textbf{0.142} & \textbf{26.67} & \textbf{0.908} & \textbf{0.187} & \textbf{33.13} & \textbf{0.861} & \textbf{0.312} & \textbf{31.63} & \textbf{0.936} & \textbf{0.098} \\
\bottomrule
\end{tabular}
}
\vspace{-1mm}
\end{table*}

\subsubsection{Effect of Loss Terms in BPR.}
We analyze the contribution of each loss term in the BPR module. 
Directly measuring the alignment between real blurry--sharp image pairs is challenging. Since alignment in the 3DGS framework is determined by camera poses, we evaluate BPR using pose errors and downstream deblurring performance.
For pose errors, as ground-truth poses are unavailable for real blurry images, we construct a synthetic evaluation set where blurry images are generated using GS-Blur and the calibrated poses of the corresponding sharp images are regard as ground-truth.
We report Translation Error (TE) and Rotation Error (RE) before and after pose optimization.
For debluring performance, we train deblurring models on data pairs generated under different loss configurations.
As shown in Table~\ref{tab:pose_loss_ablation_sup}, without pose optimization, severe misalignment leads to poor deblurring performance.
$\mathcal{L}{\mathrm{appearance}}$ mainly reduces rotation error and substantially improves deblurring performance, whereas $\mathcal{L}{\mathrm{centroid}}$ primarily reduces translation error. Their combination achieves the lowest pose errors and the best deblurring performance, demonstrating their complementarity.
\begin{table*}[t]
\centering
\caption{\textbf{Ablation on Loss Terms in BPR.}}
\vspace{-2mm}
\label{tab:pose_loss_ablation_sup}
\resizebox{\linewidth}{!}{
\begin{tabular}{c c|c|ccc|ccc|ccc|ccc}
\toprule
\multirow{2}{*}{$\mathcal{L}_{appearance}$} & \multirow{2}{*}{$\mathcal{L}_{\text{centroid}}$ } 
& \multicolumn{1}{c|}{Pose Error}
& \multicolumn{3}{c|}{RealBlur}
& \multicolumn{3}{c|}{RBVD}
& \multicolumn{3}{c|}{RSBlur}
& \multicolumn{3}{c}{BSD} \\
& & TE$\downarrow$ / RE$\downarrow$ 
& PSNR$\uparrow$ & SSIM$\uparrow$ & LPIPS$\downarrow$
& PSNR$\uparrow$ & SSIM$\uparrow$ & LPIPS$\downarrow$
& PSNR$\uparrow$ & SSIM$\uparrow$ & LPIPS$\downarrow$
& PSNR$\uparrow$ & SSIM$\uparrow$ & LPIPS$\downarrow$ \\
\midrule
\textcolor{darkred}{\usym{2717}} & \textcolor{darkred}{\usym{2717}} 
& 1.036 / 1.102 
& 27.38 & 0.880 & 0.154 & 26.46 & 0.899  & 0.189 &  32.61 & 0.854 & 0.315 & 31.42 & 0.930 & 0.109 \\
\textcolor{darkgreen}{\usym{2713}} & \textcolor{darkred}{\usym{2717}}
& 0.702 / 0.503
& 27.52 & 0.884 & 0.143 & 26.63 & 0.907 & \textbf{0.187} & 32.95 & 0.858 &  0.314 & 31.59 & 0.935 & 0.102 \\
\textcolor{darkred}{\usym{2717}} & \textcolor{darkgreen}{\usym{2713}}
& 0.452 / 0.841
& 27.43 & 0.883 & 0.146 & 26.52 & 0.903 & 0.188 & 32.75  & 0.856 & 0.315 & 31.51 & 0.934 & 0.105 \\
\textcolor{darkgreen}{\usym{2713}} & \textcolor{darkgreen}{\usym{2713}} 
& \textbf{0.445} / \textbf{0.479}
& \textbf{27.57} & \textbf{0.885} & \textbf{0.142} & \textbf{26.67} & \textbf{0.908} & \textbf{0.187} &  \textbf{33.13} & \textbf{0.861} & \textbf{0.312} & \textbf{31.63} & \textbf{0.936} & \textbf{0.098} \\
\bottomrule
\end{tabular}
}
\end{table*}

\subsubsection{Effect of CutMix-Dynamic.} 
We evaluate our designed CutMix-Dynamic on RealBlur~\cite{rim2020real} and RBVD~\cite{zhu2022deep} for camera-motion blur, and RSBlur~\cite{rim2022realistic} and BSD~\cite{zhong2020efficient} for object motion blur. Static-scene results reflect camera-blur handling, while dynamic-scene results measure robustness to object-motion blur. Results are reported in Table~\ref{tab:dynamic_ablation_sup}. 
The original CutMix~\cite{yun2019cutmix} improves performance on dynamic scenes but slightly degrades results on static scenes due to unrealistic local replacements. 
In contrast, CutMix-Dynamic introduces realistic object-motion blur and achieves the best overall performance.

\begin{table*}[t]
\centering
\caption{\textbf{Ablation on CutMix-Dynamic}}
\vspace{-2mm}
\label{tab:dynamic_ablation_sup}
\resizebox{\textwidth}{!}{
\begin{tabular}{l ccc ccc ccc ccc}
\toprule
\multicolumn{1}{c}{\multirow{2}{*}{Methods}}
& \multicolumn{3}{c}{RealBlur}
& \multicolumn{3}{c}{RBVD}
& \multicolumn{3}{c}{RSBlur}
& \multicolumn{3}{c}{BSD} \\
& PSNR$\uparrow$ & SSIM$\uparrow$ & LPIPS$\downarrow$
& PSNR$\uparrow$ & SSIM$\uparrow$ & LPIPS$\downarrow$
& PSNR$\uparrow$ & SSIM$\uparrow$ & LPIPS$\downarrow$
& PSNR$\uparrow$ & SSIM$\uparrow$ & LPIPS$\downarrow$\\
\midrule
Baseline & 27.52 & 0.884 & 0.141 & 26.60 & 0.908 & 0.186 & 33.09 & \textbf{0.860} & 0.313 & 31.52 & 0.935 & 0.100 \\
+ CutMix~\cite{yun2019cutmix} & 27.50 & 0.884 & 0.141 & 26.62 & 0.908 & \textbf{0.185} & 33.10 & \textbf{0.860} & \textbf{0.312} & 31.59 & \textbf{0.936} & 0.099 \\
+ CutMix-Dynamic & \textbf{27.58} & \textbf{0.885} & \textbf{0.140} & \textbf{26.67} & \textbf{0.909} & \textbf{0.185} & \textbf{33.12} & \textbf{0.860} & \textbf{0.312} & \textbf{31.64} & \textbf{0.936} & \textbf{0.098} \\
\bottomrule
\end{tabular}
}
\end{table*}

\subsubsection{Effect of Dataset Scale.}
The effect of dataset scale is shown in Table~\ref{tab:datascale_ablation-sup}. Performance consistently improves with increasing data scale, with the full scale achieving the best results. This trend highlights the benefit of scaling up datasets using GS-RealBlur.
\begin{table*}[t]
\centering
\caption{\textbf{Ablation on Dataset Scale}}
\vspace{-2mm}
\label{tab:datascale_ablation-sup}
\resizebox{\linewidth}{!}{
\begin{tabular}{c|ccc|ccc|ccc|ccc}
\toprule
\multirow{2}{*}{Training Data Proportions}
& \multicolumn{3}{c|}{RealBlur}
& \multicolumn{3}{c|}{RBVD}
& \multicolumn{3}{c|}{RSBlur}
& \multicolumn{3}{c}{BSD} \\
& PSNR$\uparrow$ & SSIM$\uparrow$ & LPIPS$\downarrow$ & PSNR$\uparrow$ & SSIM$\uparrow$ & LPIPS$\downarrow$ & PSNR$\uparrow$ & SSIM$\uparrow$ & LPIPS$\downarrow$ & PSNR$\uparrow$ & SSIM$\uparrow$ & LPIPS$\downarrow$\\
\midrule
25\% & 27.52 & 0.883 & 0.143 & 26.47 & 0.907 & 0.190 & 33.08 & 0.862 & 0.314 & 31.58 & 0.935 & 0.102 \\
50\% & 27.58 & 0.885 & 0.142 & 26.57 & 0.908 & 0.187 & 33.10 & 0.862 & 0.313 & 31.69 & 0.937 & 0.098 \\
100\% & \textbf{27.67} & \textbf{0.886} & \textbf{0.140} & \textbf{26.71} & \textbf{0.910} & \textbf{0.186} &  \textbf{33.15} & \textbf{0.863} & \textbf{0.311} & \textbf{31.92} & \textbf{0.939} & \textbf{0.093} \\
\bottomrule
\end{tabular}
}
\end{table*}

\section{Limitations and Future Work}
Real-world motion blur may arise from both camera motion and independently moving objects. 
Owing to the static-scene limitation of 3DGS and the still-limited reconstruction fidelity of current 4DGS methods for complex dynamic scenes, our current data construction framework primarily models camera-motion blur. 
Although CutMix-Dynamic partially compensates for this limitation by injecting real object-motion blur patterns during training, it does not explicitly reconstruct or render dynamic scenes.
In future work, advances in the reconstruction fidelity of 4D Gaussian Splatting~\cite{wu20244d,wu2026deblur4dgs} may enable our framework to be extended to dynamic scenes. 
High-speed multi-view capture could then be used to reconstruct 4D scenes and render sharp references for moving objects.

\clearpage
\clearpage

\bibliography{aaai2027}

\end{document}